\AtBeginDocument{%
  \providecommand\BibTeX{{%
    \normalfont B\kern-0.5em{\scshape i\kern-0.25em b}\kern-0.8em\TeX}}}

\documentclass[sigconf]{acmart}

\copyrightyear{2025} 
\acmYear{2025} 
\setcopyright{acmlicensed}
\acmConference[WWW Companion '25]{Companion Proceedings of the ACM Web Conference 2025}{April 28-May 2, 2025}{Sydney, NSW, Australia}
\acmBooktitle{Companion Proceedings of the ACM Web Conference 2025 (WWW Companion '25), April 28-May 2, 2025, Sydney, NSW, Australia}
\acmDOI{10.1145/3701716.3717661}
\acmISBN{979-8-4007-1331-6/2025/04}

\usepackage{graphicx}
\usepackage{dsfont}
\usepackage{subcaption}
\usepackage{colortbl}
\usepackage{balance}

\begin{document}
\newcommand{\codename}{GPT4Image}
\newcommand{\eg}{\textit{e.g.}}
\newcommand{\ie}{\textit{i.e.}}

\title{GPT4Image: Large Pre-trained Models Help Vision Models Learn Better on Perception Task}


\author{Ning Ding}
\email{dingning@stu.pku.edu.cn}
\affiliation{
  \institution{Peking University}
  \city{Beijing}
  \country{China}
}

\author{Yehui Tang}
\email{yehui.tang@huawei.com}
\affiliation{
  \institution{Huawei Noah’s Ark Lab}
  \city{Beijing}
  \country{China}
}

\author{Zhongqian Fu}
\email{fuzhongqian@huawei.com}
\affiliation{
  \institution{Huawei Noah’s Ark Lab}
  \city{Beijing}
  \country{China}
}
\author{Chao Xu}
\email{xuchao@cis.pku.edu.cn}
\affiliation{
  \institution{Peking University}
  \city{Beijing}
  \country{China}
}

\author{Kai Han}
\email{kai.han@huawei.com}
\affiliation{
  \institution{Huawei Noah’s Ark Lab}
  \city{Beijing}
  \country{China}
}

\author{Yunhe Wang}
\email{yunhe.wang@huawei.com}
\affiliation{
  \institution{Huawei Noah’s Ark Lab}
  \city{Beijing}
  \country{China}
}
\renewcommand{\shortauthors}{Ning Ding et al.} 



\begin{abstract}
The upsurge in pre-trained large models started by ChatGPT has swept across the entire deep learning community. Such powerful models demonstrate advanced generative ability and multimodal understanding capability, which quickly set new state of the arts on a variety of benchmarks. The pre-trained LLM usually plays the role as a universal AI model that can conduct various tasks like article analysis and image comprehension. However, due to the prohibitively high memory and computational cost of implementing such a large model, the conventional models (such as CNN and ViT) are still essential for many visual perception tasks. In this paper, we propose to enhance the representation ability of ordinary vision models on perception tasks (\eg~image classification) by taking advantage of the off-the-shelf large pre-trained models. We present a new learning framework, dubbed \textbf{GPT4Image}, where the knowledge  of the large pre-trained models are extracted to help CNNs and ViTs learn better representations and achieve higher performance. Firstly, we curate a high quality description set by prompting a multimodal LLM to generate descriptions for training images. Then, these detailed descriptions are fed into a pre-trained encoder to extract text embeddings that encodes the rich semantics of images. During training, text embeddings will serve as extra supervising signal and be aligned with image representations learned by vision models. The alignment process helps vision models achieve better performance with the aid of pre-trained LLMs. We conduct extensive experiments to verify the effectiveness of the proposed algorithm on various visual perception tasks for heterogeneous model architectures.
\end{abstract} 





\maketitle

\section{Introduction}
\label{sec:intro}

In the past few years, the natural language processing (NLP) community has made significant progress rapidly. Among all algorithms, the transformer model~\cite{vaswani2017attention} and self-attention mechanism laid the foundation for the success of NLP technology nowadays. At the same time, the improvement of computing power also presents opportunities for the enlargement of model size~\cite{radford2019language,touvron2023llama,smith2022using}. The advancement of large language models (LLMs) is the key factor to achieving state of the arts on various benchmarks. With the concomitant pre-training-and-fine-tuning paradigm~\cite{devlin2018bert}, LLMs can effectively handle a wild range of tasks such as sentiment analysis~\cite{sun2019utilizing}, text completion and cross-language translation~\cite{zhu2020incorporating}.

Released at the end of 2022, ChatGPT has been one of the hottest topic all around the world for the past few months, with people from all walks of life discussing and using it~\cite{lund2023chatting,biswas2023role,sallam2023chatgpt}. ChatGPT is a versatile chat robot that can provide elaborate answers after processing the user's instructs, e.g. reading an entire document to generate a comprehensive analysis report. The recently released GPT-4~\cite{openai2023gpt} model is even more impressive given its extraordinary multimodal capabilities. It's able to not only deal with diversified textual input, but also understand the content of the input image and generate detailed descriptions about it. 

However, the remarkable capabilities of GPT-4 stems from its trillions of parameters trained on thousands of GPUs. The prohibitively high computational cost makes it almost impossible for small enterprises to directly train GPT-4 level models to support their own business. In addition, the generation ability of chat robot models cannot meet the end of most application scenarios in actual production; for instance, conventional vision models such as CNNs and ViTs are still the best choices for visual perception tasks like image classification. Therefore, it's more practical and cost-effective to explore how to take advantage of pre-trained LLMs to boost the performance of existing task-specific models.

In the traditional supervised training paradigm, vision models can only learn visual representations through labels and pure pixels, while the information from other modalities is ignored. In this paper, we present a novel learning framework dubbed \codename, where the wide-range knowledge of pre-trained large models are utilized to help the conventional vision models learn enhanced representations from multi-modalities and achieve higher performance on perception task (e.g. image classification). Specifically, we first resort to the multimodal understanding ability of pre-trained LLMs through conversations to generate high-quality detailed descriptions for all images in the training dataset. Subsequently, a pre-trained text encoder is used to extract the corresponding text embeddings for all descriptions in the curated description set. After this process, the rich semantic information contained in the descriptive text generated by the LLM is compressed into a high-dimensional vector which can be easily digested by normal deep neural networks. During the training process, apart from receiving the supervising signal from ground-truth labels, the visual representations learned by the model will also be aligned with the text embeddings by minimizing a contrastive distance loss function. The modality alignment between image- and text- embeddings implicitly transfers the image comprehension knowledge of the LLM to conventional visual perception models.

The main contributions of this paper are summarized as follows: 
\begin{itemize}
	\item We propose a learning framework named \codename, where the knowledge of pre-trained LLMs are exploited to help the conventional vision models to learn enhanced representations and achieve better performance.
	\item We validate the effectiveness of the proposed algorithm on the wildly used CIFAR and ImageNet-1k datasets. Besides, we find that our method works surprisingly well on fine-grained image classification tasks (CUB200~\cite{WahCUB_200_2011} and FGVC-Aircraft~\cite{maji2013fine}) through experiments.
	\item The proposed method is consistently effective for different vision architectures, including convolutional and transformer models.
\end{itemize}


\section{Related Work}

\subsection{Large Language Models (LLMs)}

In the past few years, the natural language processing (NLP) community has made extraordinary breakthroughs since the advent of transformer~\cite{vaswani2017attention}. Language models which are based on attention mechanism such as BERT~\cite{devlin2018bert}, GPT~\cite{radford2018improving}, XLNet~\cite{yang2019xlnet}, T5~\cite{raffel2020exploring} achieved state of the arts on a variety of language benchmarks. Meanwhile, a great number of pre-training and fine-tuning algorithms were developed to further enhance the performance of transformer models. As the model size keeps increasing, researchers then discovered the data-hungry nature~\cite{hestness2017deep, oyedare2019estimating} and model scaling property~\cite{rosenfeld2021predictability, alabdulmohsin2022revisiting} of transformer architecture. These important works paved the way for the subsequent emergence of large language models, which includes LLaMa~\cite{touvron2023llama} with 65 billion parameters, GPT-3~\cite{brown2020language} with 175B parameters, BLOOM~\cite{scao2022bloom} with 176B parameters and Megatron Turing-NLG~\cite{smith2022using} with 530B parameters. These LLMs demonstrate some astonishing \textit{emergent abilities}~\cite{wei2022emergent}, such as solving math problems and analysing articles, which have never been observed on previous small language models.

\subsection{Vision-Language Pre-training (VLP)}

As both the model capacity and computational resources increase rapidly, the input of deep neural networks is no longer limited to a single modality such as image or text. Vision-language pre-training (VLP)~\cite{chen2023vlp,du2022survey} was proposed to mitigate the gap between different modalities and jointly harness the image and text cross-modality information. Benefiting from the pre-training-and-finetuning training paradigm that has prevailed in NLP, VLP models~\cite{zhang2021vinvl,zhou2020unified,kim2021vilt,lu2019vilbert,su2019vl,tan2019lxmert} quickly achieved excellent performance in various sophisticated vision-language tasks like image captioning~\cite{chen2015microsoft}, visual question answering~\cite{antol2015vqa} and visual reasoning~\cite{suhr2018corpus}. Among existing studies, one simple yet significant work is CLIP~\cite{radford2021learning}, which uses the idea of contrastive learning to seek alignment between image and text. CLIP jointly trains an image encoder and a text encoder on millions of image-text pairs collected from internet, the resulted encoder performs unexpectedly well on downstream tasks by its zero-shot classification ability.

\begin{figure*}[h]
	\centering
	\includegraphics[width=0.9\linewidth]{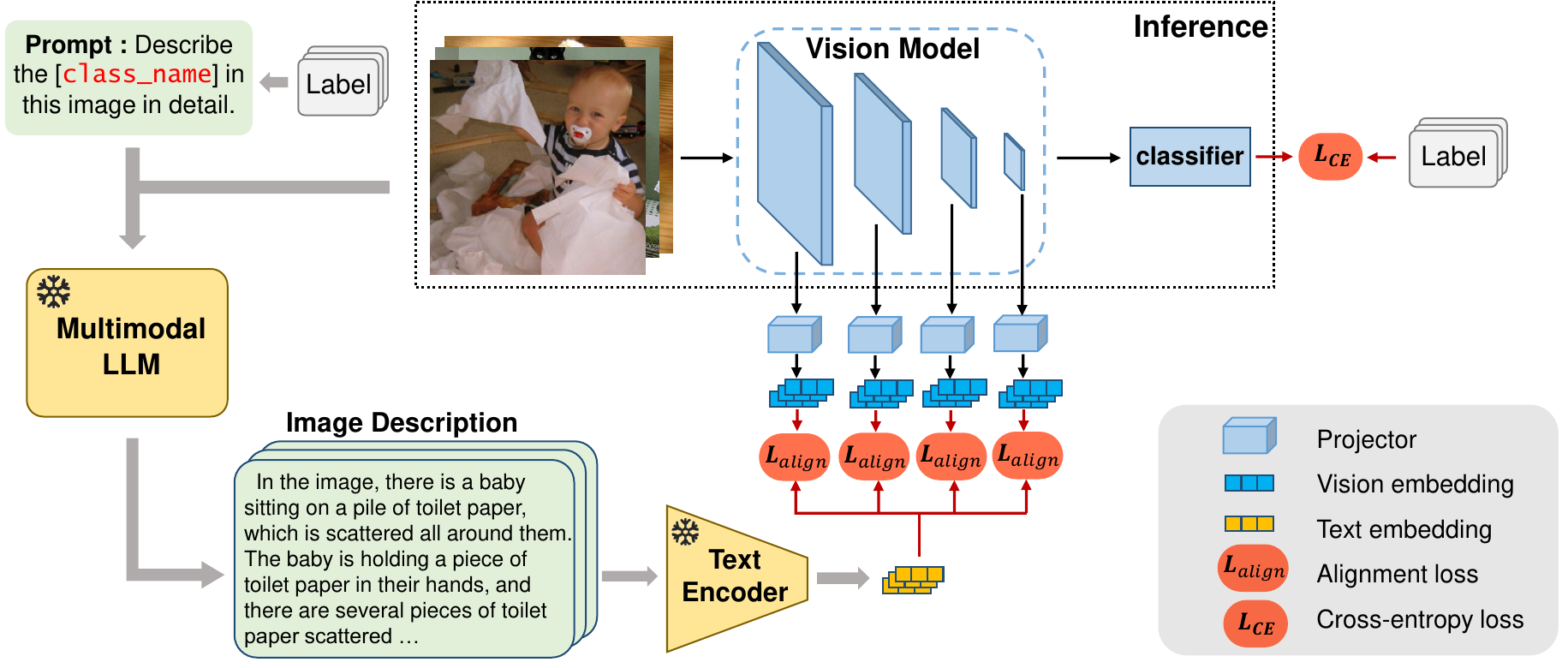}
	\caption{Overall diagram of the proposed \codename~training framework. Conventional vision models (e.g. CNN and ViT) can learn better representations with the assistance of pre-trained LLMs. Only the model within the dashed black box will be used for inference.}
	\label{fig_framework}
\end{figure*}

\subsection{Combining VLP Models with LLMs}

Compared to training the text encoder from scratch, researchers began to notice that integrating a large pre-trained generative language model directly into the visual-language system can bring significant benefits to the performance on many generative tasks, such as image captioning and visual question answering. Many works~\cite{tsimpoukelli2021multimodal,alayrac2022flamingo,chen2022visualgpt,manas2022mapl,li2023blip} utilize an off-the-shelf pre-trained LLM and freeze the its parameters to exploit the knowledge and more importantly, to save computational resources. Among all the LLMs, Flan-T5~\cite{chung2022scaling}, OPT~\cite{zhang2022opt} and GPT-2~\cite{radford2019language} are the popular encoders to be chosen due to their astonishing generative capabilities and scalabilities. One critical issue in this VLP method is how to align the image representations extracted by the visual encoder with the text embeddings output by LLM. To tackle this problem, FROZEN~\cite{tsimpoukelli2021multimodal} use the image features output by visual encoder as the prompts to instruct the LLM. VisualGPT~\cite{chen2022visualgpt} inserted a self-resurrecting activation unit into the LLM to strike a balance the language knowledge and image information. BLIP-2~\cite{li2023blip} trains a Q-Former as the adaptation layer to connect the output of visual encoder and the input of pre-trained LLM. Very recently, PaLM-E~\cite{driess2023palm} and GPT-4~\cite{openai2023gpt} with hundreds of billion of parameters demonstrate overwhelming understanding and reasoning abilities for images, languages, and even tabular data, which proves the success of such combined training paradigm.

\section{Method}

In this section, we first elaborate the overall design for the propose learning framework \codename. Then, we introduce the process of generating image descriptions by using MLLMs (multimodal LLMs). Finally, we present how to utilize these textual descriptions as supervising signals during the training phase to help vision models learn better representations and achieve higher accuracy on perception task like image classification.

\subsection{Overall Framework}

In the traditional paradigm of supervised learning~\cite{he2016deep, touvron2021training}, vision models (e.g. CNNs and ViTs) directly learn feature representations from ground truth labels and raw pixels. As recent researches have demonstrated that large pre-trained models have remarkable ability to understand and generate multimodal information, we reckon that traditional supervised learning method for visual perception tasks might as well introduce a new modality to provide diversified supervising information, instead of solely using raw pixels. 

Based on the idea of guiding the vision model to learn knowledge from different modalities, such as text, we first curate a high-quality description set to textualize the content of all images in the training dataset. Given the powerful multimodal understanding and generative abilities of large language models like GPT-4, we interact with a LLM through conversation by inputing both an training image and some prompts to make it generate a detailed description of the image. Afterwards, a pre-trained text encoder is adopted to encode the image descriptions. This operation compresses the text knowledge generated by LLM into a high-dimensional text embedding space. Finally, we minimize a contrastive distance loss function between the text embeddings and multiple image representations extracted by the target vision model to achieve cross-modality feature alignment. In addition to class labels, the descriptions generated by LLM serve as extra supervising signal of text modality to guide vision models to learning more comprehensive knowledge. Figure~\ref{fig_framework} illustrates the overall diagram of our proposed \codename~learning algorithm.

\subsection{Generating Descriptions by Prompting LLMs}
\label{method_prompt}

In this section, we explore how to generate image descriptions by resorting to LLM-based multimodal generative models, such as GPT-4~\cite{openai2023gpt}, MiniGPT-4~\cite{zhu2023minigpt} and LLaVa~\cite{liu2023llava}. These multimodal LLMs are able to take both an image and a user instruction (\textit{i.e.}~prompts) as input at the same time, and generate the corresponding answer.

By giving the prompts $$\textit{``Describe this image in detail."}$$ along with an input image, the MLLM can generate a narrative paragraph consisting of multiple sentences that elaborate the content of the image. However, the description generated in this way is likely to include both salient foreground object and unexpected background clutter in the image simultaneously. During the training process, the background information will become irrelevant semantic noise that hinders the performance of the model in vision perception tasks. This phenomenon will be demonstrated in Figure~\ref{diff_models_prompts}.

To solve this problem, we use class labels to constrain the description scope of the LLM's output. The class-conditioned prompt is formatted as $$ prompt(y_i)  =``\textit{Describe the~} {\small\texttt{[class\_name]}} \textit{~in this image in detail.}"$$ After adding the class name into the prompt, the MLLM's attention will be guided towards the desired object and the output description will contain the object's features in detail, such as shape, location, texture and so on. By using the class-conditioned prompt, we are able to create a description set for all training images. The comparison of the effects between using the two versions of prompts will be studied in Section~\ref{sec_abl_diff_prompt}.


\subsection{Learning Enhanced Image Representations}

Since we have achieved image-to-text modality transformation, in this section we present how to exploit the information of image descriptions during the supervised training process to enhance the vision model's performance.

Let $\mathbf{F}(\cdot)$ be a vision encoder model with arbitrary architecture that takes an image $x_i$ as input, and $f_i=\mathbf{F}(x_i)\in \mathds{R}^d$ be the extracted feature representation. Classification task aims to optimize model $\mathbf{F}$ and a classifier network $\hat{y_i}=\mathbf{G}(f_i)$ (\textit{e.g.} $\mathbf{G}$ is a fully-connected layer) such that the value of cross entropy loss function $L_{ce}=- \sum_i^N y_i \cdot\log \hat{y_i}$ is minimized, where $y_i$ is the corresponding label for the $i$-th image in the training dataset $\mathcal{D} = \{(x_i,y_i)\}_{i=1}^N$.

In order to enable the image descriptions to enhance the learning process of vision models, we need to extract the information contained in the text to facilitate interaction between the vision model and the MLLM. Therefore, we propose to embed the image descriptions into a $k$-dimensional feature space. We have
\begin{equation}
	f_i^{text}=\mathbf{T}(t_i)\in\mathds{R}^k
	\label{eq_text_emb}
\end{equation}
to denote the text embedding, where 
\begin{equation}
	t_i=\text{MLLM}(x_i,~prompt(y_i))
	\label{eq_desc}
\end{equation}
is the description of the $i$-th image in $\mathcal{D}$. $\mathbf{T}$ is a pre-trained text encoder such as BERT~\cite{devlin2018bert} or CLIP~\cite{radford2021learning}, and $k$ is the output embedding dimension of the encoder model $\mathbf{T}$. The text encoders generally has been pre-trained on a large-scale natural language corpus collected from Internet, and therefore has a strong capability to extract semantic information from text paragraphs, which will be investigated in Section~\ref{vis_text_emb}.
With the descriptions encoded into the embedding space, we assume that $f_i^{text}$ should be aligned with the image representation of the same sample $x_i$ so that we can maximize the use of the knowledge contained in MLLM as guidance.

Inspired by the multi-scale design adopted by many segmentation and detection architectures\cite{chen2017rethinking,ge2021yolox}, we extract multiple feature from different depth of the vision encoder $\mathbf{F}$. Taking convolutional network as an example , we use

\begin{equation}
	h_{i,s} = \mathbf{F}_{s}(x_i) \in\mathds{R}^{C_s\times W_s\times H_s}
\end{equation}
to represent the 3-D output feature map of the $s$-th stage of $\mathbf{F}$, where $\mathbf{F}_s$ denotes the partial network from the input layer to the output layer of the s-th stage, and $s \in \{1,2,...,S\}$. Then we use $S$ projectors to map these multi-scale feature maps with different resolutions to corresponding vision embeddings $f_{i,s}^{img}\in\mathds{R}^k$, where
\begin{equation}
	 f_{i,s}^{img} = \mathbf{P}_s(h_{i,s}) = \text{MLP}_s (\text{avg\_pool}(h_{i,s})) .
	\label{eq_proj}
\end{equation}

In practice, we universally set $S=4$. Each projector $\mathbf{P}_s$ consists of an MLP following an average pooling layer for convolutional neural networks and SwinTransformer~\cite{liu2021swin}. As for Vision Transformer~\cite{dosovitskiy2020image} (ViT), we equally divide ViT into 4 stages and $h_{i,s}$ denotes the feature corresponding to the {\small\textbf{\texttt{cls\_token}}} in the output token sequence of the $s$-th stage. Thus each projector $\mathbf{P}_s$ used by ViT is an MLP without average pooling layer.

To facilitate the alignment between image and text modalities, we propose to minimize a distance-based metric to pull the image representation $f_{i,s}^{img}=\mathbf{P}_s(\mathbf{F}_s(x_i))$ close to the corresponding text embedding $f_i^{text}=\mathbf{T}(t_i) $.
Concretely, the distance loss function in this work is derived from the InfoNCE loss adopted by some contrastive learning methods~\cite{he2020momentum, chen2020simple}

\begin{equation}
	L_{dist}(x_i,t_i) = \frac{1}{S} \sum_{s=1}^{S} - \log\frac{\exp(f_i^{text~\top}\cdot f_{i,s}^{img}/\tau)}{\sum_{j=1}^{|B|} \exp(f_j^{text~\top}\cdot f_{i,s}^{img}/\tau)} ,
	\label{eq_dist}
\end{equation}
where $|B|$ is the batch size and $\tau$ is the temperature hyper-parameter to control the smoothness and separability of alignment between different samples within the same mini-batch. This loss function can not only realize cross-modality matching for the same training sample, but also improve the discriminability of visual representations through contrasting among negative samples.

Therefore, the final objective function to be optimized is
\begin{equation}
	L =\sum_{i=1}^{|B|} L_{ce}(x_i,y_i) + \lambda \cdot L_{dist}(x_i, t_i) ,
	\label{eq_all_loss}
\end{equation}
where $\lambda$ is a trade-off coefficient to determine the strength of distance loss term. Based on the image descriptions generated by multimodal LLM, the supervising signals received by vision models come from not only the ground-truth labels $y_i$, but also the rich semantic information extracted from the textual descriptions. Figure~\ref{fig_framework} demonstrates the overall pipeline of the proposed learning framework.

\section{Experiments}
\label{sec_exp}

\begin{table*}[t]
	\centering
	
	\rowcolors{2}{white}{gray!20}
\caption{Top-1 Accuracy of different models on ImageNet-1K dataset.}
	\begin{tabular}{|l|c|c|}
		
	\hline
	Model &  Top-1 accuracy (\%) & Improvement (\%) \\
	\hline
	\hline
	\multicolumn{3}{|c|}{Convolutional Models} \\
	
	ResNet-50~\cite{he2016deep} & 76.3 & - \\
	ResNet-50 + \codename       & 77.8 & + 1.5 \\
	ResNet-101~\cite{he2016deep}& 77.4 & - \\
	ResNet-101 + \codename      & 79.5 & + 2.1 \\
	
	ConvNeXt-Small~\cite{liu2022convnet}& 83.1 & - \\
	ConvNeXt-Small + \codename          & 83.7 & + 0.6 \\
	ConvNeXt-Base~\cite{liu2022convnet} & 83.8 & - \\
	ConvNeXt-Base + \codename           & 84.2 & + 0.4 \\
	
	\hline
	\hline
	\multicolumn{3}{|c|}{Transformer Models} \\
	
	DeiT-Small~\cite{touvron2021training}& 79.8 & - \\
	DeiT-Small + \codename               & 80.4 & + 0.6 \\
	DeiT-Base~\cite{touvron2021training} & 81.8 & - \\
	DeiT-Base + \codename                & 82.9 & + 1.1 \\
	
	Swin-Tiny~\cite{liu2021swin}& 81.2 & - \\
	Swin-Tiny + \codename       & 81.7 & + 0.5 \\
	Swin-Base~\cite{liu2021swin}& 83.5 & - \\
	Swin-Base + \codename       & 83.9 & + 0.4 \\
	
	\hline
	\end{tabular}
	\label{tab_imagenet}
\end{table*}

\begin{table}[t]
	\centering
	\caption{Top-1 accuracy of ResNet-18 on CIFAR10 and CIFAR100 datesets.}
	\begin{tabular}{|l|c|c|}
		\hline
		\qquad\qquad Method &  Top-1 acc.(\%) & Improve (\%)\\
		\hline
		\hline
		\multicolumn{3}{|c|}{CIFAR10 dataset} \\
		ResNet-18              & 95.5 & - \\
		ResNet-18 + \codename  & 96.0 & + 0.5 \\
		\hline
		\hline
		\multicolumn{3}{|c|}{CIFAR100 dataset} \\
		ResNet-18  & 75.4 & - \\
		ResNet-18 + \codename  & 76.5 & + 1.1 \\
		\hline
	\end{tabular}
	\label{tab_cifar}
\end{table}

\subsection{Experiment on ImageNet}

\paragraph{\textbf{Datasets}} The ImageNet-1K~\cite{deng2009imagenet} dataset is a large-scale baseline benchmark for image classification task. It consists of around 1.3 M high-resolution training samples that are divided into 1,000 categories. We report the top-1 accuracy on the validation set with 50k images.

\paragraph{\textbf{Implementation details}}
In order to cover both the convolution-based and transformer-based architectures to verify the generalization ability of the proposed method, we choose four representative vision models, which are ResNet~\cite{he2016deep}, ConvNeXt~\cite{liu2022convnet}, DeiT~\cite{touvron2021training} and Swin Transformer~\cite{liu2021swin} to conduct experiments on ImageNet-1k. We train ResNet models for 120 epoch using SGD optimizer and cosine learning rate decay based on the official PyTorch example scripts\footnote{\url{https://github.com/pytorch/examples/tree/main/imagenet/main.py}}.
ConvNeXt is a modified version of convolutional network which utilizes many up-to-date components, such as large convolution kernel size, GELU activation and AdamW optimizer. We follow the official implementation\footnote{\url{https://github.com/facebookresearch/ConvNeXt}} to train ConvNeXt model for 300 epochs. Similarly, we use the same optimization and hyper-parameter settings as in their official codes for both DeiT\footnote{\url{https://github.com/facebookresearch/deit/blob/main/README\_deit.md}} and Swin Transformer\footnote{\url{https://github.com/microsoft/Swin-Transformer/tree/main}}. All transformer models are trained for 300 epochs following the official implementation.
The image descriptions are generated by MiniGPT-4~\cite{zhu2023minigpt} with class-conditioned prompt and then fed into CLIP~\cite{radford2021learning} text encoder to extract the corresponding text embeddings. The description set and all text embeddings are generated and saved before the training phase in order to reduce the high computational cost caused by repeated inference of the LLMs. During training, the text embeddings will be directly loaded into memory for reuse. We set hyper-parameters $\lambda=9$ and $\tau=0.2$ universally on ImageNet-1k.
We report the average accuracy result out of three independent runs. All the models are trained from random initialization and no extra training data is used.

\paragraph{\textbf{Experiment result}}
Table~\ref{tab_imagenet} demonstrates the experiment results on ImageNet-1K dataset. From the table we can see that the proposed \codename~method is consistently effective for vision models with different architectures.
For convolutional networks, ResNet-50 achieves 77.8\% top-1 accuracy, which is 1.5\% absolute performance growth compared to the 76.3\% accuracy of its baseline model. ResNet-101 achieves 79.5\% top-1 accuracy, which is 2.1\% absolute improvement compared to the 77.4\% baseline accuracy.
ConvNeXt-Small achieves 83.7\% top-1 accuracy, which surpasses the baseline by 0.6. ConvNeXt-Base achieves 84.2\% top-1 accuracy, which surpasses the baseline by 0.4\%.
Our proposed method further enhance the performance for transformer-based architectures. DeiT-Small attains 80.4\% top-1 accuracy, achieving 0.6\% performance improvement. And DeiT-Base attains 82.9\% top-1 accuracy, gaining 1.1\% absolute performance improvement compared to its baseline. Swin-Tiny achieves 81.7\% top1 accuracy with 0.5\% performance improvement. Swin-Base achieves 83.9\% top1 accuracy with 0.4\% performance improvement. 

\subsection{Experiment on CIFAR}

\paragraph{\textbf{Datasets.}}The CIFAR10 and CIFAR100~\cite{krizhevsky2009learning} datasets each contain 50,000 training images with 32$\times$32 low resolution that are divided into 10 categories and 100 categories, respectively. We report the top-1 accuracy on their validation set which consists of 10,000 images each.

\begin{figure}[t]
	\centering
	\includegraphics[width=1\linewidth]{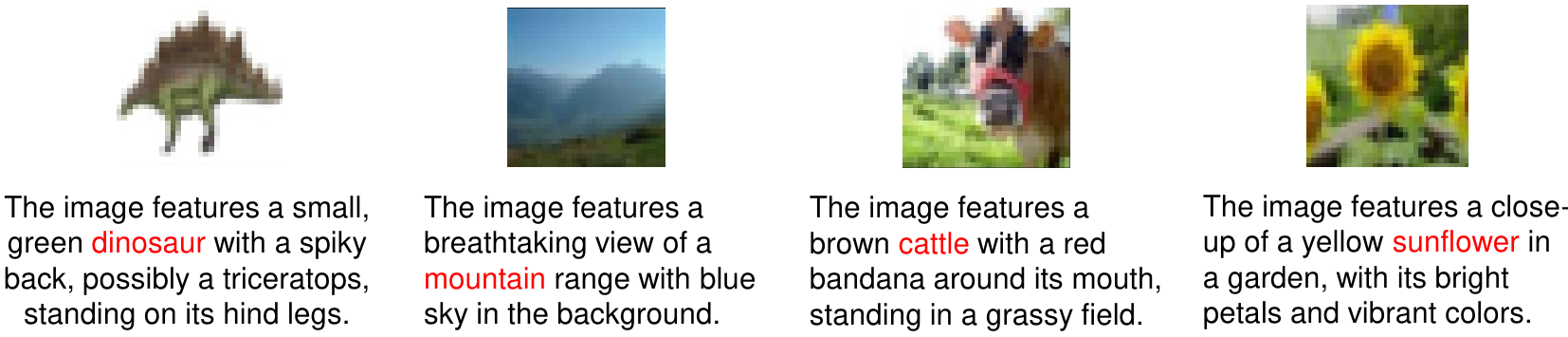}
	\caption{Examples of descriptions generated on CIFAR100. Words in color red are corresponding class names.}
	\label{fig_cifarcap}
\end{figure}

\paragraph{\textbf{Implementation details.}}
CIFAR10 and CIFAR100 are two small-scale datasets, therefore we train from scratch a light weight model ResNet-18~\cite{he2016deep} to conduct the experiments. Specifically, the model is trained with SGD optimizer for 200 epochs with batch size of 512 and weight decay of 5e-4. The initial learning rate is set to 0.1 and is decayed according to cosine schedule. We set hyper-parameters $\tau=1$ and $\lambda=0.3$ respectively for Eq.~\ref{eq_dist} and Eq.~\ref{eq_all_loss}.
We up-sample all the training images from 32$\times$32 to 224$\times$224 resolution before feeding them into the multimodal LLM for description generation. Figure~\ref{fig_cifarcap} demonstrates some examples of the descriptions generated for CIFAR-100 images by using class-conditioned prompt.

\paragraph{\textbf{Experiment result.}}
As shown in Table~\ref{tab_cifar}, the top1 classification accuracy of ResNet-18 improves by 0.5\% on CIFAR10 dataset. This is probably because the images in CIFAR10 are only divided into 10 categories and the content difference among images is relatively small. Thus large pre-trained models is unable to distinguish them well, and the generated images descriptions for the same class are semantically similar.
However, on CIFAR100 dataset, the performance improvement brought by our \codename~method is 1.1\%. Since the CIFAR100 dataset has 100 categories, with the help of class-conditioned prompt, the multimodal LLM can provide more discriminative description information for samples that belong to different categories.

\subsection{Experiment on fine-grained image classification}

\paragraph{\textbf{Datasets.}}
CUB-200~\cite{WahCUB_200_2011} dataset is a widely used fine-grained image classification benchmark that consists of 200 different bird species. The total 11,788 images are divided into a training set of 5,994 images and a testing set of 5,794 images. Figure~\ref{diff_models_prompts} demonstrates the generated description for an image from the class "cardinal".

FGVC-Aircraft~\cite{maji2013fine} dataset contains 10,000 images of aircraft spanning 100 aircraft types, most of which are airplanes. 6667 images are randomly sampled for training.
Due to the subtle inter-class differences and significant intra-class differences between subcategories, fine-grained image classification is more challenging than ordinary image classification tasks.

\paragraph{\textbf{Implementation details.}}
The image description set is also generated by MiniGPT-4~\cite{zhu2023minigpt} with class-conditioned prompt. The descriptions are then fed into CLIP~\cite{radford2021learning} text encoder to extract the corresponding text embeddings. We use ResNet-50 model on both datasets and follow the same 120-epoch training strategy as in the ImageNet-1k experiment. We performer two experimental settings, training from scratch and fine-tuning from the weights pre-trained on ImageNet-1k .

\paragraph{\textbf{Experiment result.}}
As shown in Table~\ref{tab_finegrain}, it is evident that, with the help of the proposed \codename~method, the model has made significant progress in accuracy on both fine-grained classification datasets compared to the baseline model.
Especially when trained from scratch, ResNet-50 achieves 10\% absolute accuracy improvement on CUB-200 bird classification task, and achieves 4.3\% absolute accuracy improvement on FGVC-Aircraft dataset. This indicates that the massive world knowledge stored in multimodal LLMs can help vision models more accurately identify the hard samples with very small inter-class difference and improve the overall performance in fine-grained image classification tasks.
Based on the weights pre-trained on ImageNet-1k, we further fine-tune the ResNet-50 model on these two datasets. As the last row of data in Table~\ref{tab_finegrain} shows, our method still brings sustainable growth to the classification accuracy of the model.

\begin{table}[t]
	\centering
	
	\caption{Top-1 Accuracy of ResNet-50 on CUB-200 and FGVC-Aircraft fine-grained image classification datasets.}
	\begin{tabular}{|l|c|c|}
		
		\hline
		& CUB-200 & \,FGVC-Aircraft\,\\
		\quad\qquad Method &  Top-1 acc. (\%) & Top-1 acc. (\%) \\
		\hline
		\hline
		\multicolumn{3}{|c|}{\textit{Train from scratch}} \\
		
		ResNet-50        &46.6           & 67.9 \\
		ResNet-50 + \codename&56.6~(+ 10.0\%)& 72.2~(+ 4.3\%) \\
		
		\hline
		\hline
		\multicolumn{3}{|c|}{\textit{Fine-tune from ImageNet-1k weights}} \\
		
		ResNet-50        & 78.2          & 83.4 \\
		ResNet-50 + \codename& 82.1~(+ 3.9\%)& 85.0~(+ 1.6\%) \\
		
		\hline
	\end{tabular}

	\label{tab_finegrain}

\end{table}

\begin{table*}[t]
	\centering
	\caption{Ablation study on the choice of different MLLMs and prompts. Top-1 accuracy~(\%) on ImageNet-1k is reported.}
	\begin{tabular}{|c|c|c|c|c|}
		\hline
		 & ResNet-50 & & DeiT-Base &  \\
		Method & Top-1 acc. & Improve & Top-1 acc. & Improve \\
		\hline
		\hline
		Normal Training & 76.3 &-& 81.8  &-   \\
		$t_i=\texttt{[class\_name]}_i$ & 76.5& + 0.2 & 81.9& + 0.1    \\
		\hline
		\hline
		\multicolumn{5}{|c|}{\codename~Method} \\
		\quad BLIP-2~\cite{li2023blip}\,\,\,\,\,+\,\,\, plain prompt & 76.9 & + 0.6 & 82.3& + 0.5   \\
		\,\,\,\,\, LLaVA~\cite{liu2023llava}\,\,\,\,\,+\,\,\, plain prompt & 77.2&+ 0.9&82.4& + 0.6\\
  		MiniGPT-4~\cite{zhu2023minigpt}\,+\,\,\, plain prompt & 77.4&+ 1.1&82.5& + 0.7\\
		\,\,\,\,\,\, LLaVA~\cite{liu2023llava}\,\,\, + class-conditioned prompt & 77.6 &+ 1.3 & 82.7& + 0.9 \\
		MiniGPT-4~\cite{zhu2023minigpt} + class-conditioned prompt & \textbf{77.8} & + 1.5 & \textbf{82.9}&  + 1.1 \\
		\hline
	\end{tabular}
	\label{tab_ablation_llm_prompt}
\end{table*}

\section{Ablation Study And Analysis}
\label{sec_abl}

\subsection{Effect of different prompts}
\label{sec_abl_diff_prompt}
We mentioned in Section~\ref{method_prompt} that, a properly organized prompt can better direct the MLLM's attention towards the target-class object, so that the MLLM can generate desired high-quality descriptive text.
If we input a plain prompt like \textit{``Describe this image in detail"} into the MLLM, it's likely that the MLLM would generate some irrelevant text, thus introducing some background noise to the image description. Therefore, the extracted text embedding might not have a good semantic correspondence with the image representation extracted by the vision model in the feature space. This leads to a poor embedding alignment across image- and text- modalities and results in a sub-optimal solution for the \codename~framework.
However, using a class-conditioned prompt (~\eg~\textit{``Describe the} {\small\texttt{[class\_name]}} \textit{in this image in detail."}~) can make the MLLM generates more precise and semantically rich descriptions by focusing on the target object.

\begin{figure}[t]
	\centering
	\includegraphics[width=1\linewidth]{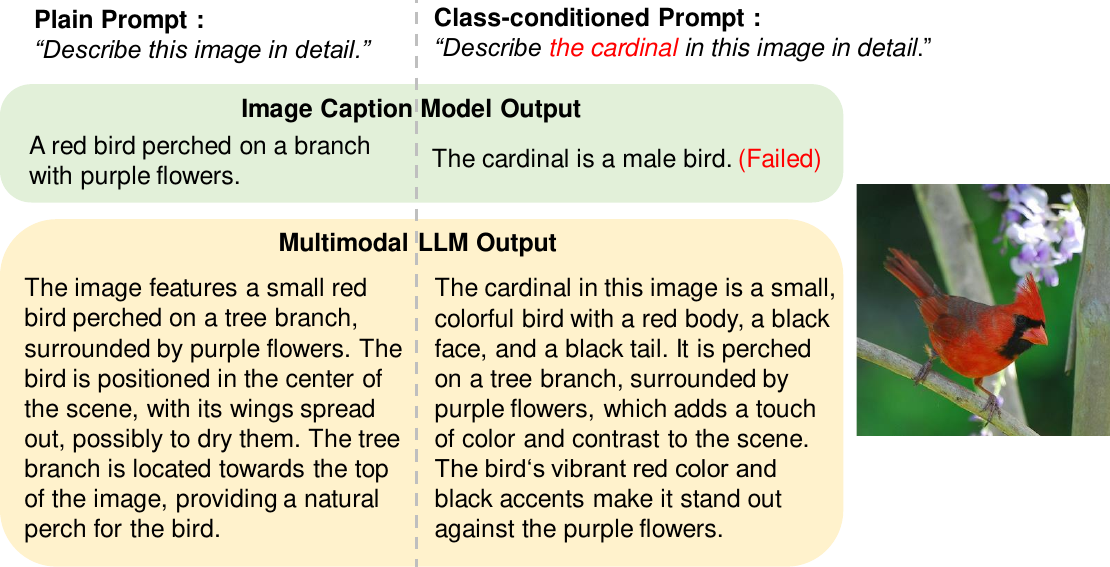}
	\caption{Image descriptions generated by different prompts and different multimodal LLMs. Information richness regarding the target object differs significantly. Green box: description from BLIP-2~\cite{li2023blip}. Yellow box: description from MiniGPT-4~\cite{zhu2023minigpt}.}
	\label{diff_models_prompts}
\end{figure}

Table~\ref{tab_ablation_llm_prompt} demonstrates the comparison results between the two prompts on ImageNet-1k. Combined with MiniGPT-4, using the class-conditioned prompt achieves a 1.5\% performance improvement~(76.3\%$\rightarrow$77.8\%)~for ResNet-50 on ImageNet-1k, while plain prompt delivers a 1.1\% performance improvement~(76.3\%$\rightarrow$77.4\%). The yellow box in Figure~\ref{diff_models_prompts} demonstrate two different descriptions generated by MiniGPT-4 when using the two types of prompts. When the input image contains multiple elements in the foreground, using {\small\texttt{[class\_name]}} as guidance could prevent the irrelevant information from appearing in the description to some extent. Especially for hard samples in fine-grained image classification tasks, the involvement of class names can stimulate the MLLM to better make use of the stored world knowledge, which increases the discriminativity of the supervising signals from the text embedding.

\begin{figure*}[t]
	\centering
	\begin{subfigure}{0.24\linewidth}
		\centering
		\includegraphics[width=0.99\linewidth]{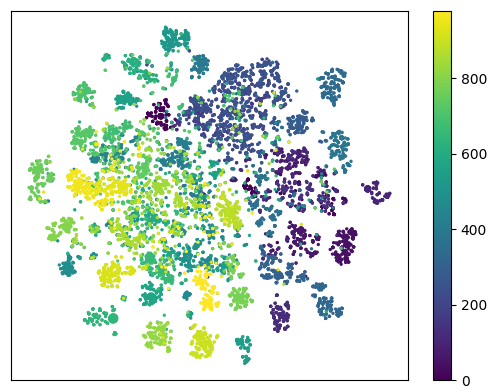}
		\caption{}
		\label{tsne_vis_short}
	\end{subfigure}
	\begin{subfigure}{0.24\linewidth}
		\centering
		\includegraphics[width=0.99\linewidth]{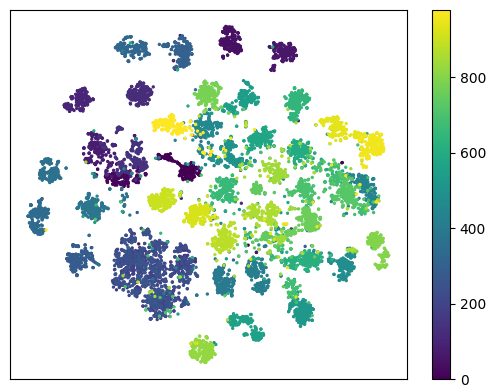}
		\caption{}
		\label{tsne_vis_long}
	\end{subfigure}
	\begin{subfigure}{0.24\linewidth}
		\centering
		\includegraphics[width=0.99\linewidth]{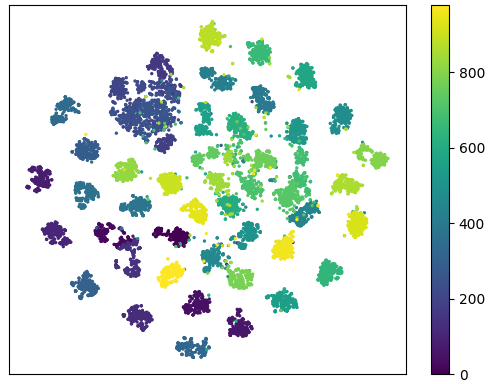}
		\caption{}
		\label{tsne_vis_clscon}
	\end{subfigure}
	\begin{subfigure}{0.24\linewidth}
		\centering
		\includegraphics[width=0.99\linewidth]{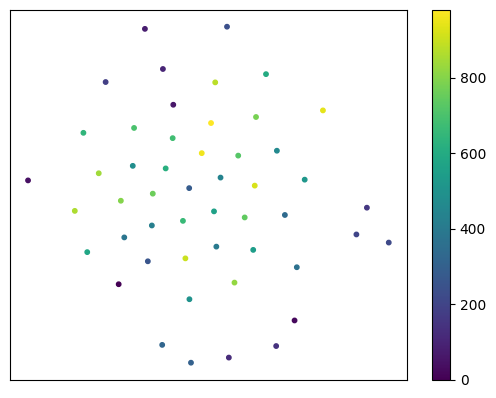}
		\caption{}
		\label{tsne_vis_clsname}
	\end{subfigure}
	\caption{t-SNE visualization of the description embeddings extracted by text encoder. Dots of the same color belong to the same category. Color bars indicates the corresponding index of the category. (a) Descriptions generated by BLIP-2 with plain prompt. (b) Descriptions generated by MiniGPT-4 with plain prompt. (c) Descriptions generated by MiniGPT-4 with class-conditioned prompt. (d) Only class names w/o using multimodal LLM. }
	\label{tsne_vis}
\end{figure*}

\subsection{Choice of Multimodal LLMs}
\label{sec_abl_diff_llm}
During the experiment, we found that using different multimodal LLMs had a significant impact on the performance improvement of the vision models. We employ BLIP-2~\cite{li2023blip}, MiniGPT-4~\cite{zhu2023minigpt} and LLaVA~\cite{liu2023llava} to conduct ablation study, and report the results in Table~\ref{tab_ablation_llm_prompt}.
As shown in the green box in Figure~\ref{diff_models_prompts}, BLIP-2 sometimes produces ambiguous sentences when dealing with the class-conditioned prompt. This is probably because BLIP-2 is specifically trained for image caption task and VQA task. The combination of BLIP-2 and plain prompt improves the top-1 accuracy on ImageNet only by 0.6\% for ResNet-50~(76.3\%$\rightarrow$76.9\%), and 0.5\% for DeiT-base~(81.8\%$\rightarrow$82.3\%).
We further compare two multimodal LLMs that are able to handle the class-conditioned prompt.
As shown in Table~\ref{tab_ablation_llm_prompt}, the combination of LLaVA and class-conditioned prompt is slightly inferior to the combination of MiniGPT-4 and class-conditioned prompt, which leads to a 1.5\% performance gain for ResNet-50~(76.3\%$\rightarrow$77.8\%) and a 1.1\% performance gain for DeiT-base~(81.8\%$\rightarrow$82.9\%).

We also create a baseline setup where we do not use the generation results of MLLM. We solely use the class names as image descriptions and directly feed them into CLIP text encoder to extract text embeddings. $t_i=\texttt{[class\_name]}_i$ in Table~\ref{tab_ablation_llm_prompt} corresponds to this baseline setup. Only a marginal accuracy improvement occurs under such setting.

\subsection{Hyper-Parameters Sensitivity}

\begin{figure}[t]
	\centering
	\includegraphics[width=1\linewidth]{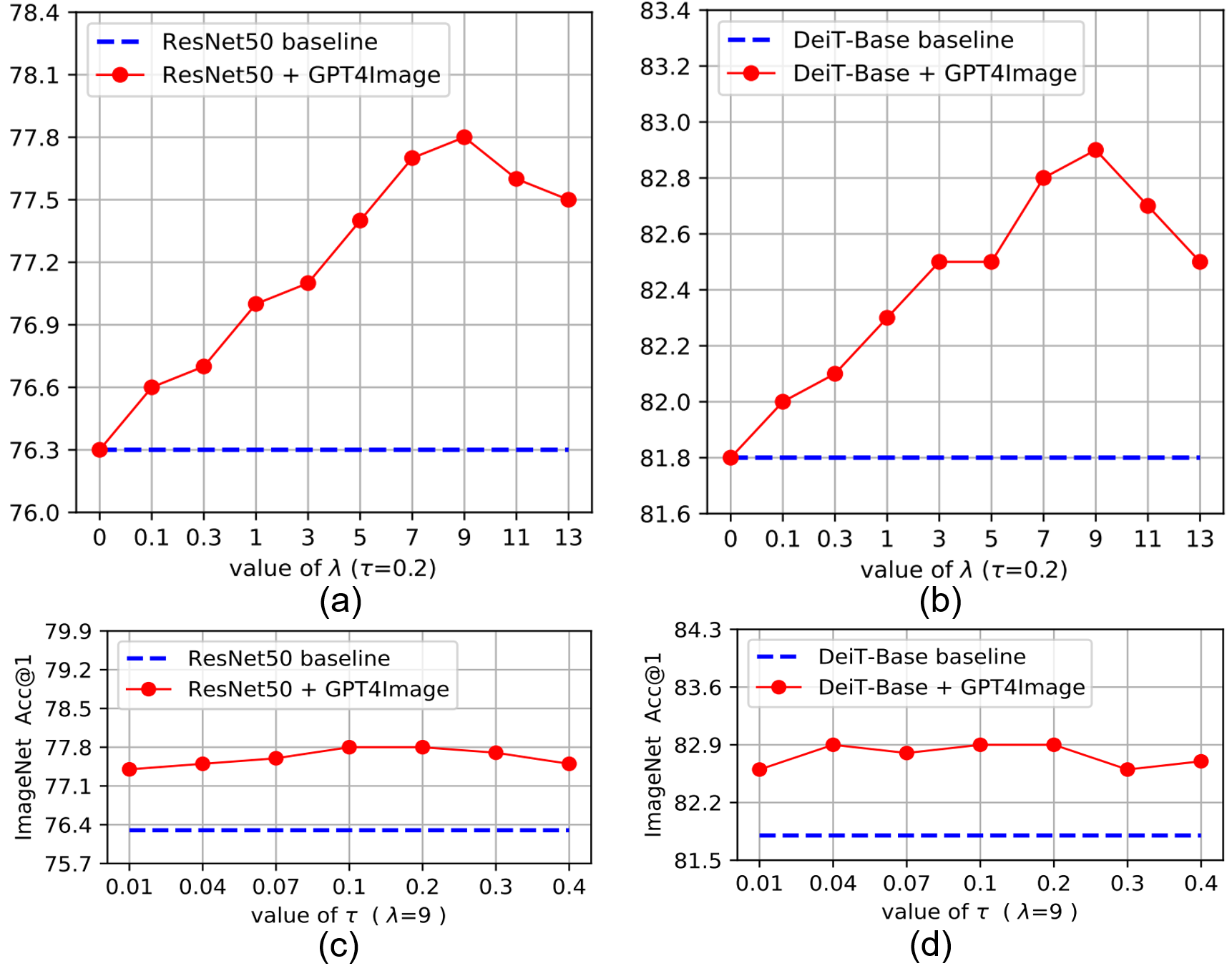}
	\caption{Hyper-parameter sensitivity analysis of $\lambda$ and $\tau$ on ImageNet-1K dataset.}
	\label{fig_hp_ana}
\end{figure}

The effect of trade-off coefficient $\lambda$ is to control the strength of the alignment loss in Eq.~\ref{eq_all_loss}. A small value of $\lambda$ is unfavorable because of insufficient utilization of the semantic information in the text embeddings.  When the value of $\lambda$ is too large, the supervising signal from the image descriptions may overtake the supervision of ground truth labels, thus leading to a sub-optimal result. 
By fixing the value of $\tau=0.2$, we find that, for both ResNet-50 and DeiT-Base, the model performance continuously grows as the value of $\lambda$ increases from 0 on the ImageNet-1k dataset. As shown in Figure~\ref{fig_hp_ana}~(a) and (b), both models obtain the highest accuracy when $\lambda=9$. The model performance starts to decrease as $\lambda$ is greater than 9.

The temperature hyper-parameter $\tau$ in Eq.~\ref{eq_dist}~controls the dispersion degree of the distribution of image representations $f_{i,s}^{img}$ in the feature space. With $\lambda$ fixed at 9, we discover that $\tau$ does not have a significant impact on model performance over a relatively wide range of values. As illustrated in Figure~\ref{fig_hp_ana}~(c) and (d), ResNet-50 and DeiT-Base consistently achieve the best performance when 
$\tau=0.1$ and $\tau=0.2$ on ImageNet-1k dataset.

\subsection{Visualization of The Text Embedding}
\label{vis_text_emb}

We visualize the text embeddings to prove that the vision models indeed receive extra supervising signals from the generated image descriptions in the presented \codename~training framework. We uniformly sample 50 classes out of the 1,000 classes of ImageNet dataset and then randomly select 250 training images from each sampled class. Then we use CLIP text encoder to extract text embeddings $f_i^{text}$ for the description $t_i$ of each image $x_i$ following Eq.\ref{eq_text_emb} and Eq.\ref{eq_desc}. We adopt t-SNE~\cite{van2008visualizing} to perform dimension reduction on $f_i^{text}$ and depict the visualization result in Figure~\ref{tsne_vis}. Dots with the same color denote the text embedding of samples that belong to the same category. The number on the color bar denotes the corresponding class index.

As illustrated in Figure~\ref{tsne_vis_long} and Figure~\ref{tsne_vis_clscon}, the natural language knowledge of pre-trained CLIP text encoder is able to map the descriptions generated by MiniGPT-4 into a benign feature space with strong discriminability. The clustering phenomenon of the text embeddings and inter-class separability property emerged in the feature space both act as guiding signals that enhance the training of vision models on perception tasks. 
Compared with the embeddings obtained by plain prompt (\ie Figure~\ref{tsne_vis_long}), the embeddings obtained by class-conditioned prompt (\ie Figure~\ref{tsne_vis_clscon}) can better distinguish the differences between categories.
As shown in Figure~\ref{tsne_vis_short}, the descriptions generated by image caption model (BLIP-2) result in text embeddings with very limited inter-class discriminability and lead to semantic ambiguity in the feature space. This hinders cross-modal alignment between the visual representation and the corresponding text embedding during training. 

We also visualize the baseline setting in Figure~\ref{tsne_vis_clsname} where we directly use the class names as descriptions to extract text embeddings. Although the embeddings corresponding to the class names have certain discriminability, there is just not enough samples to serve as supervising signals, without querying the multimodal LLM to get descriptions for all the images.
The visualization results are in agreement with the experiment results in Table~\ref{tab_ablation_llm_prompt} and the analysis of Section~\ref{sec_abl_diff_prompt}.

%

\section{Conclusion}

This paper presents a novel supervised learning framework dubbed \codename, in which conventional vision models can learn enhanced representations and achieve better performance on perception tasks (e.g. image classification) by leveraging the knowledge and multimodal ability of large pre-trained models.
Specifically, we curate a text description set for all training images by interacting with a pre-trained multimodal LLM. After extracting the embeddings of the collected descriptions with a pre-trained text encoder, we minimize a distance loss to align the text embeddings with the image representations learned by vision models. This utilizes the cross-modality knowledge of LLMs as the supervising signal to boost the training process.
We conduct extensive experiments on CIFAR and ImageNet-1K image classification benchmarks and employ both the convolutional and transformer architectures to validate the generality and effectiveness of the presented algorithm. Our method can enhance traditional vision models by taking advantage of large pre-trained models, yet without requiring extremely high computational resources to train and deploy LLMs. This approach can help small companies with limited resources benefit from LLMs and achieve higher performance in visual perception tasks. With the rapid development of large pre-trained models such as GPT-4 nowadays, the performance of conventional vision models is expected to continue improving using our proposed \codename training paradigm.

\newpage
\bibliographystyle{ACM-Reference-Format}
\balance 
\bibliography{main}

\end{document}